\documentclass{article}

\usepackage[nonatbib, final]{icbinb_neurips_2020}
\usepackage{tabularx}
\usepackage{longtable}
\usepackage{soulutf8}
\usepackage{tabularx}
\usepackage{hyperref}
\usepackage{nameref}
\usepackage{rotating}
\usepackage{geometry}
\usepackage{}
\usepackage{float}
\usepackage[nointegrals]{wasysym}
\usepackage{capt-of}
\usepackage{multirow}
\usepackage{changepage}
\usepackage{wrapfig}
\usepackage{array}
\usepackage[graphicx]{realboxes}
\usepackage{listings}
\usepackage{amsmath}
\usepackage[utf8]{inputenc} % allow utf-8 input
\usepackage[T1]{fontenc}    % use 8-bit T1 fonts
\usepackage{hyperref}       % hyperlinks
\usepackage{url}            % simple URL typesetting
\usepackage{booktabs}       % professional-quality tables
\usepackage{amsfonts}       % blackboard math symbols
\usepackage{nicefrac}       % compact symbols for 1/2, etc.
\usepackage{microtype}      % microtypography

\usepackage{wrapfig}
\usepackage{placeins}

\title{Are Gradient-based Saliency Maps Useful in Deep Reinforcement Learning?}

\author{%
    Matthias Rosynski\\
    Department of Production Engineering\\
    University of Bremen, Bremen, Germany\\
    \texttt{m\_rosynski@gmx.net}
    \And
    Frank Kirchner\\
    Department of Computer Science\\
    University of Bremen, Bremen, Germany\\
    \texttt{frank.kirchner@dfki.de}
    \And
    Matias Valdenegro-Toro\\
    German Research Center for Artificial Intelligence\\
    Bremen, Germany\\
    \texttt{matias.valdenegro@dfki.de}
}

\begin{document}

\maketitle

\begin{abstract}
    Deep Reinforcement Learning (DRL) connects the classic Reinforcement Learning algorithms with
    Deep Neural Networks. A problem in DRL is that CNNs are black-boxes and it is hard to understand the decision-making process of agents. In order to be able to use RL agents in highly dangerous environments for humans and machines, the developer needs a debugging tool to assure that the agent does what is expected.
    Currently, rewards are primarily used to interpret how well an agent is learning. However, this can lead to deceptive conclusions if the agent receives more rewards by memorizing a policy and not learning to respond to the environment. In this work, it is shown that this problem can be recognized with the help of gradient visualization techniques.
    This work brings some of the best-known visualization methods from the field of image classification to
    the area of Deep Reinforcement Learning. Furthermore, two new visualization techniques have
    been developed, one of which provides particularly good results.\\ It is being proven to what extent the
    algorithms can be used in the area of Reinforcement learning. Also, the question arises on how well the
    DRL algorithms can be visualized across different environments with varying visualization
    techniques.
\end{abstract}

\section{Introduction}
Due to the success achieved in recent years by Deep Reinforcement Learning \cite{origDQNpaper} \cite{A_Brief_Survey}, Industrial applications are becoming increasingly tangible \cite{sutton2018}. Research is being conducted on 3D map reconstruction for autonomous cars where DRL can be one of the solutions \cite{zimmermann2017learning} and also in the field of AUVs \cite{barratt2017active} as well as many other applications that interact with the real world. Once DRL algorithms are implemented on physical systems and interact with the real world, these systems can be dangerous for themselves and humans. For this reason, debugging tools are needed to understand why the agent behaves that way and whether the agent is making the right decision for the correct reason and not making a right decision for the wrong reason \cite{visNN}.

Deep reinforcement learning algorithms are nowadays interpreted and measured by the rewards the agents can get. For this reason these agents are called black box algorithms and are criticized. This makes them difficult to use in critical real-world applications.

This paper reveals that visualization techniques are a powerful debugging tool, that provides much more information than interpreting rewards. With the help of Guided Backpropagation it is even possible to locate the error in a certain layer or stream in a neural network. Moreover it is shown that Grad-Cam methods can deliver results very early in training in case of very badly trained networks. This is a big advantage especially for off-policy algorithms that take a long time to explore at the beginning of the learning process. With off-policy algorithms it can take a few days until the rewards start to increase so that the developer can determine if the neural network is learning at all \cite{Sutton_policyGradiens}.

\textbf{Contributions.} For DRL, the claim made by Adebayo et al. \cite{sanityChecks}, that Guided Backpropagation does not visualize the desired regions, but through partial input recovery it works as a kind of edge detector, is shown not to be accurate.
Also, the claim that gradient methods can be difficult to interpret, because, when answering the question "What perturbation to the input increases a particular output?", gradient methods can choose perturbations which lack physical meaning \cite{visualAtari}, is shown not always to be an issue.

Furthermore two new visualization techniques were developed, one of which provides particularly good results. These were compared and analysed with 4 other popular visualization techniques.  Their advantages and disadvantages are discussed and different fields of application for the respective visualizations are suggested.

\section{Related Work}
In this section, previous work is presented and discussed. 
As already mentioned, there is currently not a lot of work dealing with the topic of visualization in the area of deep reinforcement learning algorithms. First two works from the field of deep reinforcement learning  are presented and then and then  visualization techniques from the field of image processing.

\textbf{Semi Aggregated Markov Decision Processes.}
The authors which introduced Semi Aggregated Markov Decision Processes \cite{t-SNE} used the Atari 2600 environments as interpretable testbeds, they developed a method of approximating the behavior of deep RL policies via Semi Aggregated Markov Decision Processes (SAMDPs). They used the more interpretable SAMDPs to gain insights about the higher-level temporal structure of the policy. From a user perspective, an issue with the explanations is that they emphasize t-SNE clusters and state-action statistics which are uninformative to those without a machine learning background.

\textbf{Perturbation-based saliency methods.}
Another recently published work shows Perturbation-based saliency methods for the visualization of learned policies by Greydanus et al. \cite{visualAtari}. Their approach is to answer the question, "How much does removing information from the region around location change the policy?". The authors defined a saliency metric for image location as:
\begin{equation}
\mathcal { S } _ { \pi } ( t , i , j ) = \frac { 1 } { 2 } \left\| \pi _ { u } \left( I _ { 1 : t } \right) - \pi _ { u } \left( I _ { 1 : t } ^ { \prime } \right) \right\| ^ { 2 }
\end{equation}
The difference $\pi _ { u } \left( I _ { 1 : t } \right) - \pi _ { u } \left( I _ { 1 : t } ^ { \prime } \right)$ can be interpreted as a finite differences approximation of the directional gradient $\nabla _ { \hat { v } } \pi _ { u } \left( I _ { 1 : t } \right)$ where the directional unit vector $ \hat { v }$ denotes the gradient in the direction of $I _ { 1 : t } ^ { \prime }$.

In other words, is looking at how important were these pixels for the policy. By checking how strong the policy changes after removing some informations from the image. They use the same approach to construct Saliency Maps for the value estimate $V ^ { \pi }$ too. Greydanus et al. \cite{visualAtari} claims that gradient-based saliency methods do not yield well interpretable results. When answering the question "What perturbation to the input increases a particular output?", gradient methods can choose perturbations which lack physical meaning.

\textbf{Gradients.} The basic idea of backpropagation-based visualizations is to highlight relevant pixels by propagating the network
output back to the input image space. The intensity changes of these pixels which have the most significant impact on network decisions. Despite its simplicity, the results of saliency map are normally very noisy which makes the interpretation difficult. \cite{visBackprop}.

\textbf{Guided Backpropagation.} The idea behind guided backpropagation is that neurons act like detectors of particular image features. It is interesting in what image features the neuron detects and not in what kind of features it doesn't detect. That means when propagating the gradient, the ReLu function set all the negative gradients to zero. \cite{visBackprop} With other ways we only backpropagate positive error signals and we also restrict to only positive inputs.

\textbf{Gradient-weighted Class Activation Mapping.}
\label{sec:g_cam_meth}
One of the problems when it comes to using CAM is that a modern neural network not only consists of convolution layers but of different layers such as LSTM, MaxPooling, etc. GradCAM is an extension of CAM and it is broadly applicable to any CNN-based architectures. 
In order to obtain the class-discriminative localization map Grad-CAM $L_{\mathrm{Grad}-\mathrm{CAM}}^{c} \in \mathbb{R}^{u \times v}$ of width $u$ and height $v$ for any class $c$. First, the gradients $y^c$ of class $c$ are calculated up to the desired convolution feature map activations $A_k$. An average pooling over the width and height dimensions is then carried out \cite{gradCam}.
\begin{equation}
\alpha_{k}^{c}=\overbrace{\frac{1}{Z} \sum_{i} \sum_{j}}^{\text {global average pooling }} \underbrace{\frac{\partial y^{c}}{\partial A_{i j}^{k}}}_{\text {gradients via backprop }}
\end{equation}
Weight $\alpha^c_k$ represents a partial linearization of the deep network downstream from $A$, and captures the
"importance" of feature map $k$ for a target class $c$. On the end it  performs a weighted combination of forward activation maps with a ReLU fuction. \cite{gradCam}
\begin{equation}
L_{\mathrm{Grad}-\mathrm{CAM}}^{c}=\operatorname{Re} L U\underbrace{\left(\sum_{k} \alpha_{k}^{c} A^{k}\right)}_{\text {linear combination }}
\end{equation}
The result is a heatmap of the same size as the convolutional feature map. This feature map must then be
enlarged to the size of the original image and place it over the image to get the final result.\\
If Grad-Cam is multiplied by Guided Backpropagation with the Hadamard product we get Guided Grad-Cam \cite{gradCam}.

\textbf{Gradient Methods.} Adebayo et al. \cite{sanityChecks} deals with the informative value of visualization techniques. In this work, various visualization techniques were compared and checked whether the techniques actually depend on the model, the training data or whether it partially reconstructs the image. In other words, their work claims that Guided Backpropagation and Guided GradCam act like an edge detector, which means that they show the desired positions without showing the learned model.

In their work they randomize the weights of a model starting from the top layer, successively, all the way to the bottom layer. This procedure destroys the learned weights from the top layers to the bottom ones. Their results indicate that Guided backpropagation and Guided GradCam do not visualize the learned model, but instead partially
reconstruct the image. They interpret their findings through an analogy with edge detection in images, a technique that requires neither training data nor model.\\

\textbf{Laplacian Operator in Image Processing}. To understand one of the results we will derive the kernel of the Laplacian filter which is used for edge detection. One approach for the design of directionally invariant high-pass filters for image processing (also called edge detectors) is to use second-order derivation operators, e.g. the Laplace operator \cite{laplacian}.
For continuous functions, the Laplace operator is defined by:
\begin{equation}\mathbf{L}=\frac{\partial^{2} I[x, y]}{\partial x^{2}}+\frac{\partial^{2} I[x, y]}{\partial y^{2}}
\end{equation}
We approximate the partial derivatives by difference equations and thus obtain :
\begin{equation}\frac{\mathrm{d}^{2} \mathrm{I}[\mathrm{x}, \mathrm{y}]}{\mathrm{d} \mathrm{x}^{2}}=\{\mathrm{I}[\mathrm{x}+\Delta \mathrm{x}, \mathrm{y}]-2 \mathrm{I}[\mathrm{x}, \mathrm{y}]+\mathbf{I}[\mathrm{x}-\Delta \mathrm{x}, \mathrm{y}]\} / \Delta \mathrm{x}^{2}
\end{equation}
\begin{equation}\frac{\mathrm{d}^{2} \mathrm{I}[\mathrm{x}, \mathrm{y}]}{\mathrm{d} \mathrm{y}^{2}} \approx\{\mathrm{I}[\mathrm{x}, \mathrm{y}+\Delta \mathrm{y}]-2 \mathrm{I}[\mathrm{x}, \mathrm{y}]+\mathrm{I}[\mathrm{x}, \mathrm{y}-\Delta \mathrm{y}]\} / \Delta \mathrm{y}^{2}
\end{equation}
Thus with $\Delta x, \Delta y=1$ (except for the sign) the mask for the Laplace operator is :
\begin{equation}\mathbf{L}_{1}=\left[\begin{array}{rrr}
0 & -1 & 0 \\
-1 & 4 & -1 \\
0 & -1 & 0\end{array}\right]
\label{L1}
\end{equation}
Numerous other approximations of the Lapace operator are possible. Examples are
the following masks (their transfer functions have the form known from the Gaussian and binomial distributions)\cite{laplacian} 
\begin{equation}\mathbf{L}_{2}=\left[\begin{array}{ccc}
0 & -1 & -1 \\
-1 & 8 & -1 \\
-1 & -1 & 0
\end{array}\right] \quad \mathbf{L}_{3}=\left[\begin{array}{ccc}
1 & -2 & 1 \\
-2 & 4 & -2 \\
1 & -2 & 1
\end{array}\right] \quad \mathbf{L}_{4}=\left[\begin{array}{ccc}
-1 & -2 & -1 \\
-2 & 12 & -2 \\
-1 & -2 & -1
\end{array}\right]
\label{L3}
\end{equation}

\section{Experimental Setup}

Experiments were performed in two different environments a simple (Breakout-v0) and a complex one (Seaquest-v0) from OpenAI Gym \cite{brockman2016openai}. Four different agents were implemented. DDDQN (4 frames as input) \cite{sewak2019deep}, Splitted Attention DDDRQN (with LSTM), and two on Policy gradient algorithms A3C (3 frames as input) \cite{a3cOrigin} and an A3C Agent with LSTM.

The Splitted Attention DDDRQN was by far the best trained agent and receives most of the points (Seaquest-v0 9521 Points) that is why most of the results are referring to this agent \cite{attention}.

In order to better examine the visualization methods and to enable better interpretation, the original frames of the games are placed over the gradients with a $50\%$ opacity. In this way it is possible to assign the gradients to the features and to better interpret the results.

With the off policy algorithms not only the output is visualized but also the Q-value and the Advantage stream. To proof the results of Ziyu Wang et al. \cite{dueling}. In order to maintain comparability between the visualization techniques, the same state was always visualized in this work. This also applies to the visualization of the actor and critic agents, as well as the visualization of the value and advantage streams.
Two new visualization techniques have been developed and the following visualization methods have been implemented. Some features of the implementation are discussed below.

\textbf{Gradient and Guided Backpropagation.}
Compared to image processing, where the gradients of the guided backpropagation or gradient method are normalized over the image, the gradients in a video can also be normalized over the entire video.
For this reason, both have been implemented and tested. First, the gradients output were normalized for each state. Then the gradient and guided back propagation method around the whole video was normalized and tested.\\
In the visualizations, the gradients were checked in relation to different output layers. In particular, this work examines the advantage and the value stream in the off-policy algorithms in more detail, as well as the last layer that delivered the best results. In the actor critic methods both outputs of the NN were examined.

\textbf{Grad-Cam and Guided Grad-Cam.}
\label{sec:con_grad_cam}
With the Grad-Cam and Guided Grad-Cam method, the visualization can be applied to different convolutional layers. An example of what the differences look like is also shown. The best results were achieved on the first convolutional layer across all agents and across all environments. For this reason, the results of the Grad-Cam methods are usually applied to the first convolutional layer.\\\newline
\textbf{G1Grad-Cam and G2Grad-Cam.}
Two new visualization techniques were developed and will be presented in this work, which are also compared. Both visualization techniques are a further development of Grad-Cam and Guided Grad-Cam. The idea behind it is the same as for guided back propagation, that neurons act like detectors of particular image features. And we are interested in what image features the neuron detects and not in what kind of features it does not detect. That means when propagating the gradient, we set all the negative gradients to 0.
In the further course, Grad-Cam with the Guided Model is called G1Grad-Cam and GradCam with the Guided Model multiplied by Guided backpropagation is called G2Grad-Cam.\\

\section{Experimental Results}
Since a state usually contains several frames, the results of the gradient methods (unless other-wise stated) are related to the last frame in the sequence. With the Grad-Cam methods the results (unless otherwise stated) relate to the first convolutional layer. Furthermore, the word \emph{stable} is defined in the context of visualization techniques in this paper as follows: gradients or visual highlights that can be seen on most frames in a video. Gradients or visual highlights that can be seen on every 4th, 5th (or even less) frame in a video are called \emph{not stable}.

First we discuss the visualization techniques, then we look closer at the class discriminative visualization algorithms and finally we make a hypothesis about the importance of negative gradients in backpropagation algorithms. Detailed results are all provided in the appendix (Figures \ref{fig:dddqn_r_grad} to \ref{fig:a3cLSTM_c_grad_cam}).

\subsection{Visualization Algorithms}
\textbf{Gradient.}
Gradients method delivers a lot of noise which often mixed with the important features. It is only very poorly suited for debugging neural networks in the deep reinforcement learning area. We get similar results with the agent in all environments. This can be seen in Figures: \ref{fig:dddqn_r_grad}, \ref{fig:dddqn_s_grad_a}, \ref{fig:dddqn_s_grad_v}, \ref{fig:spatt_grad}, \ref{fig:spatt_grad_diver}, \ref{fig:back_swa}, and \ref{fig:break_a3c_crit_grad}. 

\textbf{Guided Backpropagation.}
Guided backpropagation delivers the best and most stable results under all environments and among all agents.
Also, that it shows negative gradients is a very good indication of how well an agent is trained. It can be seen that the better the agent has been trained, the stronger the transition between negative and positive gradients.

Another interesting observation that guided back propagation (compared to the Grad-Cam methods) did not visualize the agent so well during the breakout game. Even in the DDDQN network, where the agent was well trained, only weak gradients could be recognized on the agent himself on Figures \ref{fig:dddqn_r_a} and \ref{fig:dddqn_r_guid_back}. In contrast, on Seaquest, which is a much more complex environment and the agent was not well trained, very strong gradients were displayed on the agent itself, as  Figures \ref{fig:dddqn_s_grad_a} and \ref{fig:dddqn_s_grad_v} show.

Another comparison that has been made in this work is how the visualization of the guided backpropagation algorithm changes with respect to normalization. In other words, how does the visualization change when the normalization of the gradients is carried out over a frame compared to when the normalization was created over the entire video.

The results showed that when the frame was normalized, the gradients in the video flashed more strongly. With normalization across the entire video, the transitions from the frames were smoother. In this way, the less known / visited states can be recognized because the gradients on which are weaker to recognize.
This method is also the most suitable for visualizing the differences between the advantage and the value stream in the off-policy algorithms. Guided back propagation was able to identify most of the features in any environment and these were also very stable across the video.

During the development of the split attention DDDQN agent, the guided back propagation method for debugging was also used. An interesting finding that was made during development was that although the neural network had several errors, results were still very good (over 3000-3500 points, similar results were achieved by the normal attention DDDRQN network). Only after the Advantage and Value Stream was visualized did it become apparent that the network only learned on the Advantage side and that the gradients on the value side were very strong but chaotic. Based on the points, the errors that were made during programming would not have been noticed. Altogether two errors could be discovered and they could even be localized in the value stream. One mistake was incorrectly linking the layer and another when merging the two streams.

When developing the A3C with LSTM network, guided back propagation was also used to debug the network. Before a working A3C with LSTM was developed in this work, attempts were first made to train the agent with bidirectional layers. This was unsuccessful. In the visualization, no gradients were displayed when the gradients from the output layer to the input layer were calculated (the output gradients during printing were also 0). Then the network with guided backpropagation was examined more precisely with this visualization technique, because in contrast to the Grad-Cam methods, the gradient methods calculate the gradients from each individual layer to the input, this means that the individual layers can be examined. It turned out that the agent started to learn in the first three layers and no longer from the bidirectional layers. When these layers were replaced by LSTMs, the agent immediately began to learn even in the higher layers.

\textbf{Grad-Cam.}
This method also showed stable results (based on the first convolutional layer), even if the results were often issued in inventory, especially with less well-trained agents, as seen in Figure \ref{fig:guided_grad325}.

Looking at the higher layers, some features could also be visualized, but here the visualization was inverted more often, as shown in Figure \ref{fig:grad_cam2_inverted}, and the specified position of the features in the visualization algorithm was less precise, shown in Figures \ref{fig:grad_cam2} and \ref{fig:grad_cam3}.\\
One advantage over the guided backpropagation method which is noticeable, is that with less well trained agents, the visualization (even if mostly often inventoried) was able to visualize the features faster and without interference. However the G1Grad-Cam could beat these results.

Grad-Cam was able to achieve better results in the Breakout environment (see Figure: \ref{fig:dddqn_r_grad_cam}), where the agent could be visualized relatively poorly with guided backpropagation. Another advantage of the grad cam method (at least for neural networks without LSTM) compared to guided backpropagation is that, as can be seen in Figure \ref{fig:dddqn_r_grad_cam}, the time flow can be visualized. Means how important the previous frames were.

Due to the negative gradients that exist in guided backpropagation, a superimposition or averaging of the gradients from all input frames would not provide a similar result, since the gradients would cancel each other out with guided backpropagation.

Since the Grad-Cam has a ReLU function, only positive results are displayed. However, this has the disadvantage that we cannot differentiate as well if an agent is well trained. The visualization differences are minimal as can be seen in the Actor Critic between Figures \ref{fig:a3c_act_grad_cam} and \ref{fig:a3c_crit_grad_cam} and different between the Splitted Attention DDDRQN (See Figure \ref{fig:spatt_grad_cam}) and the DDDQN Agent . However, the results with Grad-Cam are better  with poorly trained agents and have less disruption than with guided backpropagation. One reason for this could be that the grad-cam carried an average pooling across the height and width of the dimension are carried out. With average pooling, the tendencies could be displayed better and there would be fewer disturbances due to averaging. This could be the reason why by Grad-Cam by weakly trained networks achieve better results than with guided backpropagation.

A disadvantage compared to guided back propagation, which can be seen in all agents is that not all features can be visualized. By the breakout environment  the area of the image where the agent breaks through the last line and receives many points could be not visualized (see Figure \ref{fig:dddqn_r_guid_back} and Figure \ref{fig:dddqn_r_grad_cam}). In Seaquest, Grad-Cam was unable to visualize the oxygen bar, nor the divers collected.

\textbf{Guided Grad-Cam and G2Grad-Cam.}
The combination of guided back propagation and one of the Grad-Cam methods has proven to be very unstable. One reason is that not all features are displayed in the Grad-Cam method, as has already been mentioned. But the main reason is that the results are shown often inverted by the Grad-Cam method. Due to the inversion, the areas with the features have a value of 0, which means that the multiplication with the guided back propagation also results in 0.

Basically, with these two methods, no additional knowledge can be obtained that has already been obtained with guided back propagation or Grad-cam.

\textbf{G1Grad-Cam}
The G1Grad-cam method gave much better results for the DDDQN agent in the game breakout and also the time flow than with the Grad-Cam method, as shown in Figure \ref{fig:dddqn_r_g1grad_cam} (G1Grad-Cam) versus Figure \ref{fig:dddqn_r_grad_cam} (Grad-Cam). Interestingly, the G1Grad-Cam method has problems visualizing the features in game Seaquest.

The G1Grad-Cam method shows less inverted results than the Grad-Cam method. However, especially with less well-trained agents, the results are not stable or not exactly in position. Basically, the method only convinced the DDDQN agent in the game Breakout. Here G1grad-Cam was able to display the ball and the agent as well as their past positions very clearly than all other visualization techniques.

However, the upper breakthrough area could not be visualized as the guided back propagation method did.\\\newline
An interesting aspect of the poorly trained splitted attention agent is that in Seaquest the G1Grad-Cam method gives the best interpretable results over all visualization methods, as shown in Figure \ref{fig:g1grad_cam325} (in contrast to the well trained one, where it gives no results). The neural network was saved after 325 episodes (for comparison, the well-trained neural network was saved after 5300 episodes).  After 325 episodes the neural network has done about 250 000 steps and 75\% of them have chosen a random action (because of exploration). After this short period of time, the G1Grad-Cam method was able to deliver well interpretable results.

For comparison: After 1500 episodes it can be determined via the rewards that the neural network is learning. This means that with the G1Grad-Cam method it is possible to recognize up to 5 times faster if the neural network is learning at all. This makes it a helpful debugging tool.

\subsection{Action Discriminative Visualization Algorithms}
The fact that the agent has learned to recognize features does not mean that the agent has learned how to behave. The agent recognizes e.g. the oxygen bar and understand when it ends that the episode will end, but it has not learned what to do about it. It is similar with the fishes that the agent recognizes them but does not necessarily mean that the agent will avoid the fish or knows what to do with it. For this reason, action discriminative algorithms were examined more closely.

The investigations did not give any indication that Grad-Cam and G1Grad-Cam, which are action discriminative algorithms, have a visible visual relationship between the state and the action taken. The expectation that Grad-Cam would emphasize a particular fish because the agent would swim away from a fish or target a fish, could not be confirmed.

\subsection{Hypothesis about Negative Gradients in Guided Backpropagation and their Meaning}
Apparently negative gradients also play a role. With the DDDQN agent, the negative gradients were time-dependent. The newest frame has the most positive and the oldest most negative gradients and the negative gradients are always shown where the ball was or will be. One reason for this could be the different architecture of the neural network which only calculates the error of the selected action but not the error of the non-selected actions compared to the other developed CNN. With the other agents, it can be observed that the negative gradients did not form in a time-dependent but position-dependent manner. The strong negative gradients have formed around the features. It can also be seen that the stronger the positive and negative gradients, the better the agent.

\begin{wrapfigure}{r}{0.3\textwidth}
    \centering
    \vspace*{-1em}
    \includegraphics[width=0.3\textwidth]{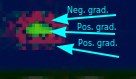}
    \caption{The agent is covered with positive gradients followed by negative and positve again.}
    \label{fig:lap}
\end{wrapfigure}

A possible explanation could be that with guided backpropagation the gradients resemble a matrix or the kernel of an edge detector. And the Laplacian Operator has strong positive and negative values in the matrix to detect edges. Backpropagation could create a matrix that has learned to better identify features with strong negative and positive values, just like with a Laplace operator. Figure \ref{fig:lap} shows that strong positive gradients are followed by strongly negative gradients and some small positve again (compare in Equations \ref{L1} and \ref{L3}, Equation $L_1$ and $L_3$). The Laplace operator has a similar structure. The difference is that the Laplace operator detects edges and not more complex features. Guided backpropagation would create a kind of Laplace matrix for features based on this hypothesis. This could be a possible explanation for the role of negative gradients.

\section{Conclusions and Future Work}

\textbf{Visualisation techniques}. First of all, the assumptions made in the paper "Visualizing and Understanding Atari Agents" \cite{visualAtari}, that guided backpropagation cannot be used for visualization techniques and the assertion made in the paper "Sanity Checks for Saliency Maps" \cite{sanityChecks}, that guided backpropagation and guided Grad-Cam (at least in image processing) do not visualize the learned model but work similarly to an edge detector do not seem apply to deep reinforcement learning policies.

Visualization methods are very well suited as an additional, if not main debugging tool. Especially Guided Backpropagation was used during the development of Splitted Attention DDDQN with bidirectional LSTM agents and the A3C with LSTM agent to detect various errors in the neural network. Not only can errors be detected but also an assignment could be made in which stream the error was located and which layer caused problems. This kind of debugging is not possible with the usual reward evaluation. Especially since the faulty neural network could achieve better results than the original one despite the errors it contains, these errors would not have been noticed.

Furthermore, it was also shown with two agents (Figure \ref{fig:a3c_crit_guid_back} and Figure \ref{fig:dddqn_s_guid_back_v}) that Guided backpropagation can be used to detect if the agent has learned to avoid an area. The agent obtained a high reward, but this is not the expected behavior. The agent has learned to stay in the water at a certain height. The agent always shoots to the left or to the right if something comes up there. This is an important insight because it means that if the environment changes the agent will probably not be able to cope with these changes as well as an agent that has a lower score, but swims specifically towards the fishes. Here a simple reward function could even lead to misinterpretations. Guided backpropagation could show the different strength of the gradients on the fish that swam at different heights, which could lead to the conclusion that the agent does not visit certain areas of the environment.

The Grad-Cam and especially the G1Grad-Cam method achieves very early results. This is useful during the training process of the neural network. For example, if an agent has to be trained for 14 days in a HPC cluster, G1Grad-Cam can quickly provide initial information after a few minutes / hours as to whether the agent recognizes any features at all or whether there are indications of errors in the neural network. The rewards that are currently used as the main debugging tool are hardly usable at the beginning, especially with off-policy algorithms. Because the agent is exploring at the beginning, the rewards are random, but the neural network learns to recognize the features even if it does not yet know what they mean or how the agent has to behave. This is a big time advantage over the Reward function that took 2 days with the Split Attention Agent to show the first signs that the Rewards are increasing. 

The error does not necessarily have to be in the neural network, but also a faulty implementations can be detected in this way (during this work wrong stacking of the frames could be visualized on the neural network and the error could be corrected by the DDDQN agent). 
The G1Grad-Cam and Grad-Cam method can save time when designing neural networks during debugging compared to a pure interpretation by the Reward function.

However, no evidence was found that using G1/Grad-Cam methods which are action discriminative methods, can visualize the difference in importance between the features, that has the biggest impact on the action (agent's decision). No feature was highlighted that is more important for the decision of an action.

Guided backpropagation was able to recognize most of the features. All features detected by the Grad-Cam methods and more. Even though the gradients are sometimes hardly visible and only after long training the DDDQN agent could visualize itself in the game Breakout.\\
This method is especially suitable to evaluate the agent at the end (Does the agent recognize all important features? Are there differences in intensity between the same features in different states? How strong are the gradients on the features?) or for debugging purposes. Guided backpropagation is better suited for debugging than the Grade-Cam method, because individual streams could be examined and individual layers. In this way the error can be identified more quickly under certain circumstances (as in the development of the Split Attention Agent).

\textbf{Negative gradients.} A hypothesis about possible explanation for the importance of negative gradients that forms around features were put forward in this work. The explanation was compared to a Laplacian filter which revealed the edges. However, edges were not revealed here, but entire features it was concluded that through backpropagation a kind of Laplacian kernel feature detector was visualized. Where the negative gradients would be produced by a second derivative to better detect the features.

\textbf{Future work}. After having shown in this work that visualization techniques are a basic important tool for debugging neural networks, there are still some important questions left. 
First of all the features that are important for the agent could be visualized. What could not be shown in this paper is that in action discriminative methods, the features that have a greater importance for the selected action are more strongly emphasized. A possible reason for this could be that by averaging all feature maps of the Grad-Cam methods the differences between the normal features and the features that directly influenced the agent's decision were very small. Here, a special implementation of Guided backpropagation might help to better identify the differences, by reprogramming the Guided backpropagation to an action discriminative algorithm.  This would give a better understanding of the advantages shown in this work, that guided backpropagation can better evaluate the intensity of how well a neural network recognizes the feature. In combination with a modified action discriminative Guided backpropagation algorithm, it might be possible to get more meaningful results if it is possible to visualize the feature that had the greatest influence on the current action.\\

\clearpage
\bibliographystyle{plain}
\bibliography{references}

\clearpage
\appendix

\section{Detailed List of Experiments}
The agent and the environment are shown on the left and the visualization techniques on the right.\\

\begin{table}[H]
\small 
\centering
\begin{adjustwidth}{-2.2cm}{}
\begin{tabular}{|p{2.5cm}|c||c|c|c|c|c|c|c|}\hline
& & \multicolumn{2}{c|}{Q-Values}&\multicolumn{4}{c|}{1st conv. Layer}\\
Agent & Env. & Gradient & Guid. back. & Grad-C. & Guid. Grad-C. & G1Grad-C. & G2Grad-C. \\\hline\hline
DDDQN \cite{sewak2019deep} & Breakout-v0 & \checked  & \checked  & \checked  & \checked  & \checked  & \checked  \\
DDDQN \cite{sewak2019deep} & Seaquest-v0 & \checked  & \checked  & \checked  & \checked  & \checked  & \checked  \\\hline
Split. At. DDDQN (bi. LSTM)& Seaquest-v0  & \checked  & \checked  & \checked  & \checked  & \checked  & \checked  \\\hline
A3C \cite{a3cOrigin} & Breakout-v0 & \checked  & \checked  & \checked  & \checked  & \checked  & \checked  \\
A3C \cite{a3cOrigin} &Seaquest-v0 & \checked  & \checked  & \checked  & \checked  & \checked  & \checked  \\\hline
A3C with LSTM \cite{a3cOrigin}& Seaquest-v0 & \checked  & \checked  & \checked  & \checked  & \checked  & \checked  \\\hline
\end{tabular}
\end{adjustwidth}
\caption{Combinations of visual explanation experiments}
\end{table}$\;$\newline
In addition, there are the different settings of the visualization techniques, e.g. which layer or stream was visualized with gradient methods or up to which convolutional layers were the Grad-Cam method applied:\\

\begin{table}[H]
\small 
\centering
\begin{adjustwidth}{-1.5cm}{}

\begin{tabular}{|p{2.5cm}|c||c|c|c|c|c|}\hline
 &  & Val. Stream& Adv. Stream &1st conv. L.&2st conv. L.&3st conv. L.\\
Agent & Env. & \multicolumn{2}{c|}{Guided backpropagation} & \multicolumn{3}{c|}{Grad-Cam}\\\hline\hline
DDDQN \cite{sewak2019deep} & Breakout-v0 &  \checked  & \checked  & \checked  &  &   \\\hline
Split. At. DDDQN (bi. LSTM)& Seaquest-v0  & \checked  & \checked  & \checked & \checked & \checked \\\hline

\end{tabular}
\end{adjustwidth}
\caption{Combinations of different settings for visual explanation experiments}
\end{table}$\;$\newline
The past frames in the input of the neural network for the guided backpropagation algorithm are also examined:
\begin{table}[H]
\small 
\centering
%\begin{adjustwidth}{-1.5cm}{}

\begin{tabular}{|l|c||c|c|c|c|}\hline
Agent & Env.& t-1 & t-2 &t-3 &t-9\\\hline\hline
DDDQN \cite{sewak2019deep} & Breakout-v0 &  \checked  & \checked  & \checked  &  \\\hline
Split. At. DDDQN (bi. LSTM)& Seaquest-v0  &   &   &  & \checked \\\hline

\end{tabular}
%\end{adjustwidth}
\caption{Combinations of time-dependent visual explanation experiments}
\end{table}$\;$\newline
An other question also arises: When does the neural network show first signs of feature recognition? Is it possible to find out faster if the neural network or the agent in general has been programmed correctly? Than about the interpretation of the reward function, which is only possible after a long time because of the exploration of the agent in off-policy algorithms.\\ 
For this reason, a neural network that was after only 325 episodes / 250 000 steps saved, is also tested (for comparison: the well trained agent had trained over 5300 episodes and about 13 000 000 steps):\\
\begin{table}[H]
\begin{adjustwidth}{-2.2cm}{}
\small 
\centering
\begin{tabular}{|p{2.0cm}|c||c|c|c|c|c|c|}\hline
Agent & Env. & Gradient & Guid. back. & Grad-C. & Guid. Grad-C. & G1Grad-C. & G2Grad-C. \\\hline
Split. At. DDDQN (bi. LSTM)& Seaquest-v0  &  \checked & \checked  & \checked & \checked & \checked & \checked \\\hline

\end{tabular}
%\end{adjustwidth}
\caption{Combinations of time-dependent visual explanation experiments}
\end{adjustwidth}
\end{table}

\FloatBarrier
\section{Detailed Experimental Results
}
The visualized results are presented in red or green color.  Meanwhile red is used for negative gradients, green has been used for positive ones. Since the Grad-Cam methods have a ReLu function, the results are always positive. All Images are taken at the same time for better comparison.

\FloatBarrier
\subsection{DDDQN: Breakout-v0}

%------------------------------------------- DDDQN: Breakout-v0 ------------------------------------------
%------DDDQN--INTRO-------
\begin{figure}[!hb]
	\begin{minipage}[t]{0.45\textwidth}
		\includegraphics[width=\textwidth]{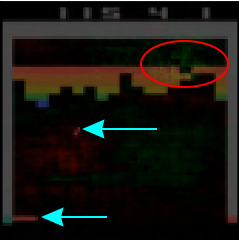}
		
		\vspace*{-0.25cm}
\caption{Gradient visualization method. In the red oval you can see that the NN-Agent started to understand which region gives him higher future rewards and where it has to shot. This is because if the agent breaks through the last line, the ball will bounce off the box and the ceiling and get a lot of reward.}
\label{fig:dddqn_r_grad}
	\end{minipage}
	\hfill
	\begin{minipage}[t]{0.45\textwidth}
		\includegraphics[width=\textwidth]{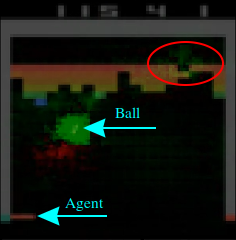}
		
		\vspace*{-0.25cm}
\caption{Guided backpropagation visualization method. The ball is with this method well heighlighted. In the red oval you can see that the NN-Agent started to understand which region gives him higher rewards and where it has to shot}
\label{fig:dddqn_r_guid_back}
	\end{minipage}
\end{figure}
\begin{figure}[H]
		\begin{minipage}[t]{0.45\textwidth}
		\includegraphics[width=\textwidth]{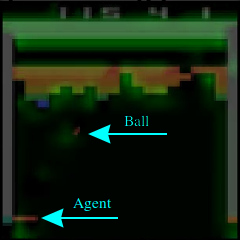}
		
		\vspace*{-0.25cm}
\caption{Grad-Cam. Some visual highlights on the agent as well as on the ball are clearly visible. Furthermore you can see the direction the ball is coming from as all 4 frames are visualized with the Grad-Cam method.}
\label{fig:dddqn_r_grad_cam}
	\end{minipage}
	\hfill
	\begin{minipage}[t]{0.45\textwidth}
		\includegraphics[width=\textwidth]{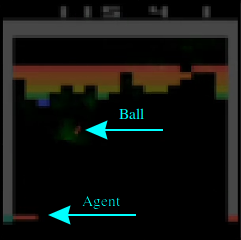}
		\vspace*{-0.25cm}
\caption{Guided Grad-Cam. Since Guided Grade-Cam is a multiplication of the Grade-Cam method and backpropagation, it is also obvious that we only visualize sub areas of the original source. The negatives and positive gradients of the current frame calculated by the guided backpropagation method are multiplied by the degree-cam method, which does its job on all frames. And only if both have positive high values this area will be highlighted.}
\label{fig:dddqn_r_guided_grad_cam}
	\end{minipage}
\end{figure}
\begin{figure}[H]
		\begin{minipage}[t]{0.45\textwidth}
		\includegraphics[width=\textwidth]{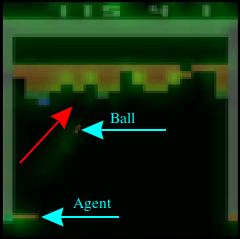}
		
		\vspace*{-0.25cm}
\caption{G1Grad-Cam. Comaprad to the Grad-Cam method we can see much more stable visualized features. We can see clearly the old positions of the ball and when the agent moves the old position of the Agent because we have four frames as input. This is different to the Gradient methods where we cant ovalep the positive and negative gradients, thats why we are using the gradient methods on the last frame.}
\label{fig:dddqn_r_g1grad_cam}
	\end{minipage}
	\hfill
	\begin{minipage}[t]{0.45\textwidth}
		\includegraphics[width=\textwidth]{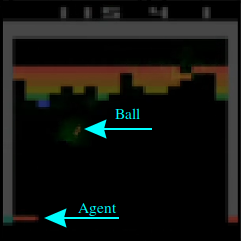}
		
		\vspace*{-0.25cm}
\caption{G2Grad-Cam. Since Guided Grade-Cam is a multiplication of the Grade-Cam method and backpropagation, it is also obvious that we only visualize sub areas of the original source. The negatives and positive gradients of the current frame calculated by the guided backpropagation method are multiplied by the degree-cam method, which does its job on all frames. And only if both have positive high values this area will be highlighted.}
\label{fig:dddqn_r_g2grad_cam}
	\end{minipage}
\end{figure}
%------------------------

\begin{figure}[H]
	\begin{minipage}[t]{0.45\textwidth}
		\includegraphics[width=\textwidth]{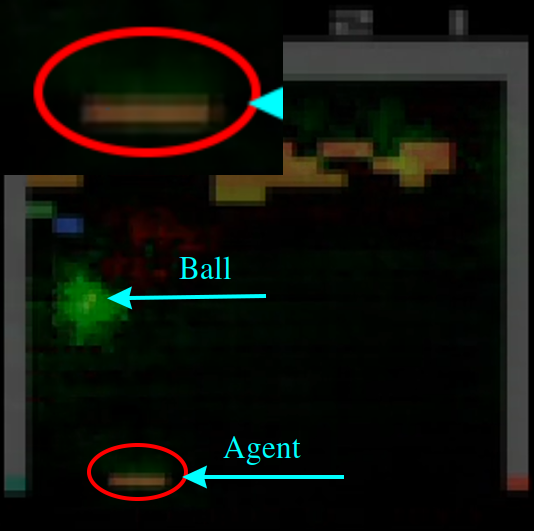}
		
		\vspace*{-0.25cm}
\caption{Guided backpropagation (advantage stream). In the advatage straem we can see that the NN is slightly more focusing on his own position than in the value stream}
\label{fig:dddqn_r_a}
	\end{minipage}
	\hfill
	\begin{minipage}[t]{0.45\textwidth}
		\includegraphics[width=\textwidth]{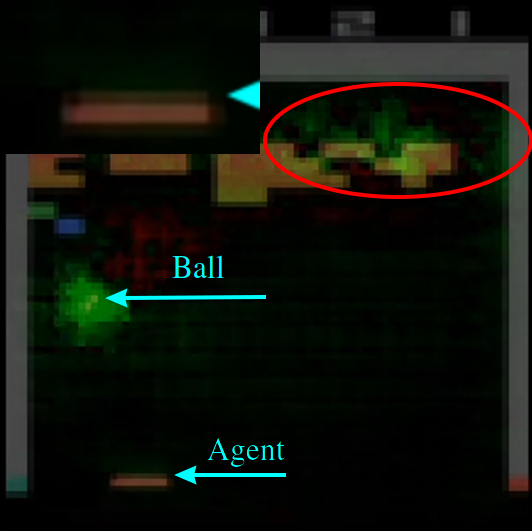}
		
		\vspace*{-0.25cm}
\caption{Guided backpropagation (value stream)
. In the value stream we see that the gradients are stronger on focused on the future rewards. This is because if the agent breaks through the last line, the ball will bounce off the box and the ceiling and get a lot of reward.}
\label{fig:dddqn_r_v}
	\end{minipage}

\end{figure}
%--------------------t-x---------
\begin{figure}[H]
	\begin{minipage}[t]{0.45\textwidth}
		\includegraphics[width=\textwidth]{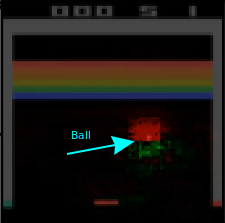}
		
		\vspace*{-0.25cm}
\caption{Guided backpropagation (t-1). We can see here that the old position of the ball (frame t-1) has positive gradients, whereas the current position of the ball is very much wrapped in negative gradients.}
\label{fig:t-1}
	\end{minipage}
	\hfill
	\begin{minipage}[t]{0.45\textwidth}
		\includegraphics[width=\textwidth]{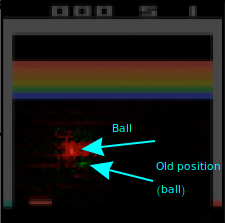}
		
		\vspace*{-0.25cm}
\caption{Guided backpropagation (t-2). We can see here that the old position of the ball (frame t-2) has positive gradients, whereas the current position of the ball is very much wrapped in negative gradients.}
\label{fig:t-2}
	\end{minipage}
\end{figure}
\begin{figure}[H]
	\begin{minipage}[t]{0.45\textwidth}
		\includegraphics[width=\textwidth]{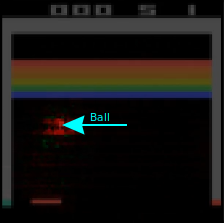}
		\caption{Guided backpropagation (t-3). We can see here that the old position of the ball (frame t-3) has positive gradients, whereas the current position of the ball is very much wrapped in negative gradients.}
\label{fig:t-3}
\end{minipage}
\end{figure}
%------------------------------------------ DDDQN: Seaquest-v0 ------------------------------------------

\subsection{DDDQN: Seaquest-v0}

\begin{figure}[H]
	\begin{minipage}[t]{0.45\textwidth}
		\includegraphics[width=\textwidth]{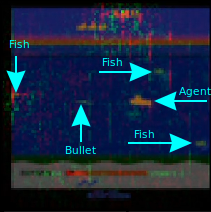}
		
		\vspace*{-0.25cm}
\caption{Gradient (advantage stream). We can see on the Agent some gradients more than on the value Stream, but in general very noisy results.}
\label{fig:dddqn_s_grad_a}
	\end{minipage}
	\hfill
	\begin{minipage}[t]{0.45\textwidth}
		\includegraphics[width=\textwidth]{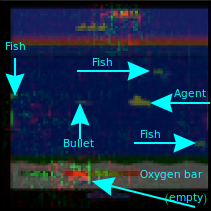}
		
		\vspace*{-0.25cm}
\caption{Gradient (value stream). We can see on the Oxygen bar some gradients more than on the advantage Stream, but in general very noisy results.}
\label{fig:dddqn_s_grad_v}
	\end{minipage}
\end{figure}
\begin{figure}[H]
	\begin{minipage}[t]{0.45\textwidth}
		\includegraphics[width=\textwidth]{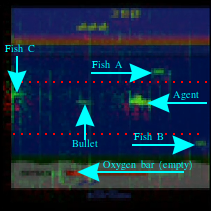}
		
		\vspace*{-0.25cm}
\caption{Guided backpropagation (advantage stream). In this picture we can see that the NN envelops more gradients around fish C than on fish B or A which have hardly any gradients. This is because the agent has learned that if it stays on the height between the red lines and only shoots to the left or right when something comes up it survives longer and pays no attention to the rest of the environment. The agent has learned his behaviour by heart and would not be able to react as well as an agent who makes less points but reacts better to the environment. If the environment were to change slightly (e.g. the fish would suddenly swim from the top right to the bottom left) the agent could hardly react to it because it ignores everything that is not in the area between the lines.}
\label{fig:dddqn_s_guid_back_a}
	\end{minipage}
	\hfill
	\begin{minipage}[t]{0.45\textwidth}
		\includegraphics[width=\textwidth]{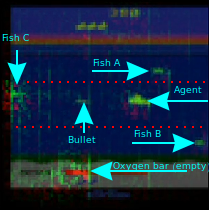}
		
		\vspace*{-0.25cm}
\caption{Guided backpropagation (value stream). In the Value Stream there is a small difference to the advatage stream. We see that the agent has no oxygen and has to appear. The DDDQN agent has never learned how to surfaced in seaquest, however he realises that the oxygen bar is an important feature that is related to the end of the episode, but the agent has not learned how to do what. And since the agent has learned to leave the area between the lines, he will not be able to learn how to do it. }
\label{fig:dddqn_s_guid_back_v}
	\end{minipage}
\end{figure}
\begin{figure}[H]
	\begin{minipage}[t]{0.45\textwidth}
		\includegraphics[width=\textwidth]{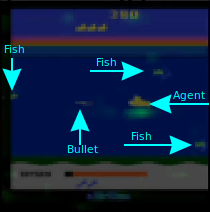}
		
		\vspace*{-0.25cm}
\caption{Grad-Cam}
\label{fig:dddqn_r_a}
	\end{minipage}
	\hfill
	\begin{minipage}[t]{0.45\textwidth}
		\includegraphics[width=\textwidth]{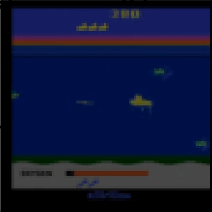}
		
		\vspace*{-0.25cm}
\caption{Guided Grad-Cam}
\label{fig:dddqn_r_v}
	\end{minipage}
\end{figure}

%--------------------------------------------------
\begin{figure}[H]
	\begin{minipage}[t]{0.45\textwidth}
		\includegraphics[width=\textwidth]{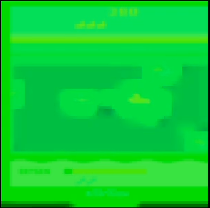}
		
		\vspace*{-0.25cm}
\caption{G1Grad-Cam. Compared to Breakout-v0, the DDDQN agent did not give good results for Seaquest-v0. }
\label{fig:dddqn_r_a}
	\end{minipage}
	\hfill
	\begin{minipage}[t]{0.45\textwidth}
		\includegraphics[width=\textwidth]{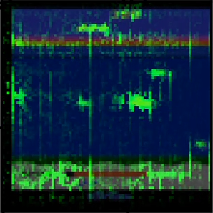}
		
		\vspace*{-0.25cm}
\caption{G2Grad-Cam}
\label{fig:dddqn_r_v}
	\end{minipage}
\end{figure}

\subsection{Split At. DDDQN: Seaquest-v0}

%----------------------------------------- Split At. DDDQN: Seaquest-v0 ----------------------------------------
%-------------------------------Seaquest DDDQN--------------

\begin{figure}[H]
	\begin{minipage}[t]{0.45\textwidth}
		\includegraphics[width=\textwidth]{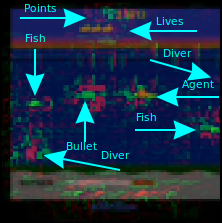}
		
		\vspace*{-0.25cm}
\caption{Gradient. Very noisy results.}
\label{fig:spatt_grad}
	\end{minipage}
	\hfill
	\begin{minipage}[t]{0.45\textwidth}
		\includegraphics[width=\textwidth]{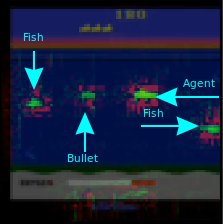}
		
		\vspace*{-0.25cm}
\caption{Guided backpropagation. Since this agent has a very well trained NN network and has achieved the best results in the game, you can can see here very clearly how  the positive gradients on the features are surrounded by negative gradients.}
\label{fig:spatt_guid_back}
	\end{minipage}
\end{figure}
\begin{figure}[H]
	\begin{minipage}[t]{0.45\textwidth}
		\includegraphics[width=\textwidth]{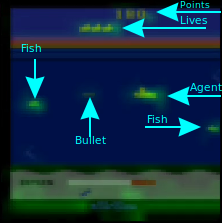}
		
		\vspace*{-0.25cm}
\caption{Grad-Cam. Good visible highlights of the most important features.}
\label{fig:spatt_grad_cam}
	\end{minipage}
	\hfill
	\begin{minipage}[t]{0.45\textwidth}
		\includegraphics[width=\textwidth]{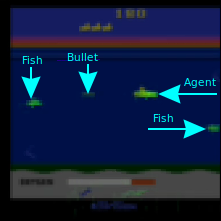}
		
		\vspace*{-0.25cm}
\caption{Guided Grad-Cam}
\label{fig:spatt_guid_grad_cam}
	\end{minipage}
\end{figure}
\begin{figure}[H]
	\begin{minipage}[t]{0.45\textwidth}
		\includegraphics[width=\textwidth]{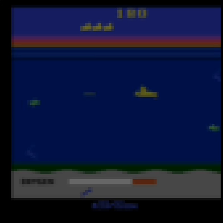}
		
		\vspace*{-0.25cm}
\caption{G1Grad-Cam. No visible highlights over the whole video.}
\label{fig:spatt_g1grad_cam}
	\end{minipage}
	\hfill
	\begin{minipage}[t]{0.45\textwidth}
		\includegraphics[width=\textwidth]{pictures/results/splitted_attention/intro/g1grad_cam.png}
		
		\vspace*{-0.25cm}
\caption{G2Grad-Cam.  No visible highlights over the whole video.}
\label{fig:spatt_g2grad_cam}
	\end{minipage}
\end{figure}
\begin{figure}[H]
	\begin{minipage}[t]{0.45\textwidth}
		\includegraphics[width=\textwidth]{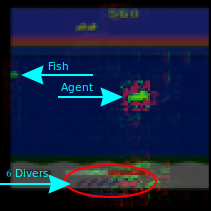}
		
		\vspace*{-0.25cm}
\caption{Guided backpropagation (advantage stream)}
\label{fig:spatt_a}
	\end{minipage}
	\hfill
	\begin{minipage}[t]{0.45\textwidth}
		\includegraphics[width=\textwidth]{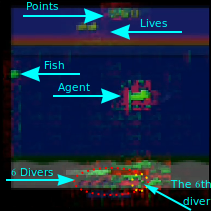}
		
		\vspace*{-0.25cm}
\caption{Guided backpropagation (value stream). The agent has learned to pay more attention to the collected divers in the Value Stream. Here he has collected all 6 divers, which means that when he shows up he will get extra reward. Furthermore you can see that the fish in the value stream has more gradients than in the advantage stream.}
\label{fig:spatt_v}
	\end{minipage}
\end{figure}

\begin{figure}[H]
	\begin{minipage}[t]{0.45\textwidth}
		\includegraphics[width=\textwidth]{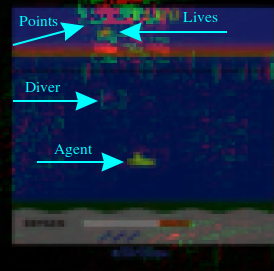}
		
\caption{Gradient. In this frame we see that the agent also recognizes the divers that contain a long term reward when the agent collects 6 divers and brings them to the surface.}
\label{fig:spatt_grad_diver}
	\end{minipage}
	\hfill
	\begin{minipage}[t]{0.45\textwidth}
		\includegraphics[width=\textwidth]{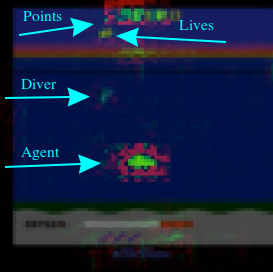}
		
\caption{Guided backpropagation. In this frame we see that the agent also recognizes the divers that contain a long term reward when the agent collects 6 divers and brings them to the surface.}
\label{fig:spatt_guid_back_diver}
	\end{minipage}
\end{figure}

%--------------------------------------------------

\begin{figure}[H]
	\begin{minipage}[t]{0.45\textwidth}
		\includegraphics[width=\textwidth]{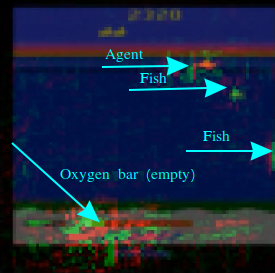}
		
\vspace*{-0.30cm}
\caption{Gradient. Here we see that the agent with the gradient method also pays attention to the empty oxygen bar. This is also the reason why the agent appeared.}
\label{fig:back_swa}
	\end{minipage}
	\hfill
	\begin{minipage}[t]{0.45\textwidth}
		\includegraphics[width=\textwidth]{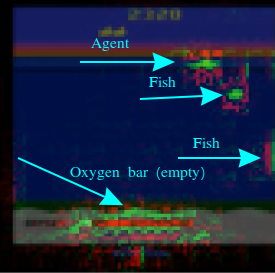}
		
\vspace*{-0.30cm}	
\caption{Guided backpropagation. In this figure we can see that the agent with the gradient method also pays attention to the empty oxygen bar - in contrast to the gradient method the oxygen bar is completely wrapped in gradients. }
\label{fig:gui_back_swa}
	\end{minipage}

\end{figure}

%---------------------GRAD-CAM-----------------------------
\begin{figure}[H]
	\begin{minipage}[t]{0.45\textwidth}
		\includegraphics[width=\textwidth]{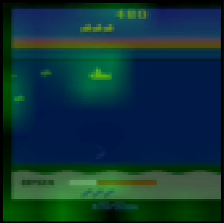}
		
		\vspace*{-0.30cm}
\caption{Grad-Cam 2nd convolutional layer. When we visualise the second layer, with the very well trained NN we always see no connections that we can interpret but not as well as we saw with the first layer}
\label{fig:grad_cam2}
	\end{minipage}
	\hfill
	\begin{minipage}[t]{0.45\textwidth}
		\includegraphics[width=\textwidth]{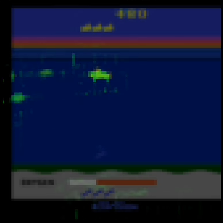}
		
		\vspace*{-0.30cm}
\caption{Guided Grad-Cam 2nd convolutional layer}
\label{fig:guided_grad_cam2}
	\end{minipage}
\end{figure}

\begin{figure}[H]
	\begin{minipage}[t]{0.43\textwidth}
		\includegraphics[width=\textwidth]{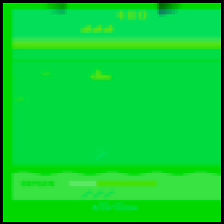}
		
		\vspace*{-0.30cm}
\caption{G1Grad-Cam 2nd convolutional layer. As in the first layer, this method does not show well interpretable results}
\label{fig:g1grad_cam2}
	\end{minipage}
	\hfill
	\begin{minipage}[t]{0.45\textwidth}
		\includegraphics[width=\textwidth]{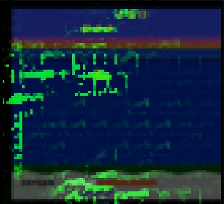}
		
		\vspace*{-0.30cm}
\caption{G2Grad-Cam 2nd convolutional layer}
\label{fig:g2grad_cam2}
	\end{minipage}
\end{figure}
%---------------------------3rd conv.layer----------------------
\begin{figure}[H]
	\begin{minipage}[t]{0.44\textwidth}
		\includegraphics[width=\textwidth]{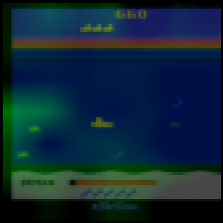}
		
		\vspace*{-0.30cm}
\caption{Grad-Cam 3rd convolutional layer. Hardly interpretable results. Partially invented highlights of the features.}
\label{fig:grad_cam3}
	\end{minipage}
	\hfill
	\begin{minipage}[t]{0.44\textwidth}
		\includegraphics[width=\textwidth]{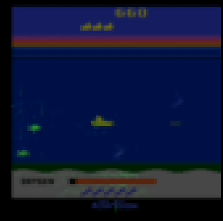}
		
		\vspace*{-0.30cm}
\caption{Guided Grad-Cam 3rd convolutional layer}
\label{fig:guided_grad_cam3}
	\end{minipage}
\end{figure}

\begin{figure}[H]
	\begin{minipage}[t]{0.44\textwidth}
		\includegraphics[width=\textwidth]{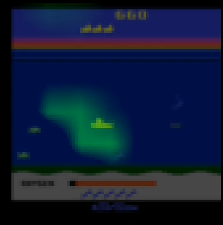}
		
		\vspace*{-0.30cm}
\caption{G1Grad-Cam 3rd convolutional layer. Better results in terms of agent position than the Grad-Cam method.}
\label{fig:g1grad_cam3a}
	\end{minipage}
	\hfill
	\begin{minipage}[t]{0.44\textwidth}
		\includegraphics[width=\textwidth]{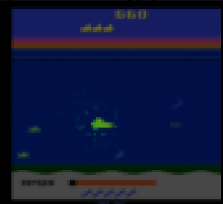}
		
		\vspace*{-0.30cm}
\caption{G2Grad-Cam 3rd convolutional layer }
\label{fig:g2grad_cam3}
	\end{minipage}
\end{figure}

\begin{figure}[H]
	\begin{minipage}[t]{0.45\textwidth}
		\includegraphics[width=\textwidth]{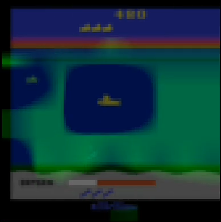}
		
		\vspace*{-0.30cm}
\caption{Grad-Cam 2nd convolution layer (inverted gradients). In the 3rd layer we can often observe inverted gradients with the Grad-Cam method as shown in this figure.}
\label{fig:grad_cam2_inverted}
	\end{minipage}
\hfill
	\begin{minipage}[t]{0.45\textwidth}
		\includegraphics[width=\textwidth]{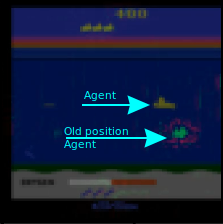}
		
		\vspace*{-0.30cm}
\caption{Guided backpropagation t-9. In this layer we see the visualization of the frame t-9 projected on the current frame.}
\label{fig:t-9}
	\end{minipage}
\end{figure}
%----------------------------------- not well trained agent--
\begin{figure}[H]
	\begin{minipage}[t]{0.45\textwidth}
		\includegraphics[width=\textwidth]{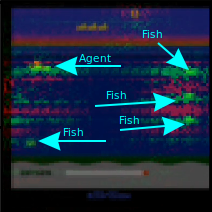}
		
		\vspace*{-0.30cm}
\caption{Gradient; episodes: 325. First gradients form around the features. Although the agent still plays randomly and does not know what these features mean, he begins to understand that they have an influence on the decisions of the agent. There is little difference between the gradient method and Guided backpropagation.}
\label{fig:grad325}
	\end{minipage}
	\hfill
	\begin{minipage}[t]{0.45\textwidth}
		\includegraphics[width=\textwidth]{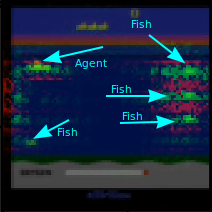}
		
		\vspace*{-0.30cm}
\caption{Guided backpropagation; episodes: 325. First gradients form around the features. Although the agent still plays randomly and does not know what these features mean, he begins to understand that they have an influence on the decisions of the agent. There is little difference between the gradient method and Guided backpropagation.}
\label{fig:guided_back325}
	\end{minipage}
\end{figure}

\begin{figure}[H]
	\begin{minipage}[t]{0.45\textwidth}
		\includegraphics[width=\textwidth]{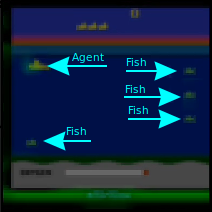}
		
		\vspace*{-0.30cm}
\caption{Grad-Cam; episodes: 325}
\label{fig:grad_cam325}
	\end{minipage}
	\hfill
	\begin{minipage}[t]{0.45\textwidth}
		\includegraphics[width=\textwidth]{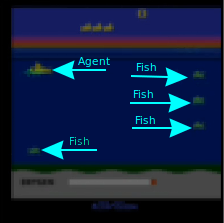}
		
		\vspace*{-0.30cm}
\caption{Guided Grad-Cam; episodes: 325. The Grad-Cam method gives clearly better and more interpretable results than the gradient methods.}
\label{fig:guided_grad325}
	\end{minipage}
\end{figure}

\begin{figure}[H]
	\begin{minipage}[t]{0.45\textwidth}
		\includegraphics[width=\textwidth]{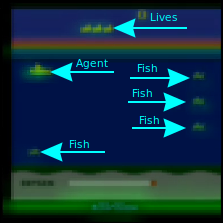}
		
		\vspace*{-0.30cm}
\caption{G1Grad-Cam; episodes: 325. The G1Grad-Cam method gives the best results we can on the fish and the agent and also on the number of lives of the agent see some highlights.}
\label{fig:g1grad_cam325}
	\end{minipage}
	\hfill
	\begin{minipage}[t]{0.45\textwidth}
		\includegraphics[width=\textwidth]{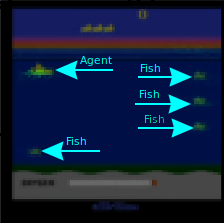}
		
		\vspace*{-0.30cm}
\caption{G2Grad-Cam; episodes: 325}
\label{fig:g2grad_cam325}
	\end{minipage}
\end{figure}

%--------------------------------------------- A3C: Breakout-v0 --------------------------------------------

\subsection{A3C: Breakout-v0}

\begin{figure}[H]
	\begin{minipage}[t]{0.45\textwidth}
		\includegraphics[width=\textwidth]{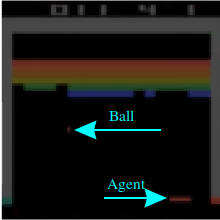}
		
				\vspace*{-0.30cm}
\caption{Actor: Gradient. Gredients so small that you can't see them without a strong multiplication factor.}
\label{fig:break_a3c_act_grad}
	\end{minipage}
	\hfill
	\begin{minipage}[t]{0.45\textwidth}
	   \includegraphics[width=\textwidth]{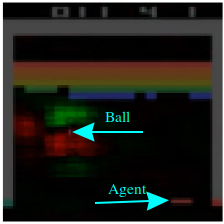}
	   
	   		\vspace*{-0.30cm}
\caption{Critc: Gradient. You can see some arbitrary gradients around the ball.}
\label{fig:break_a3c_crit_grad}
	\end{minipage}
\end{figure}
\begin{figure}[H]
	\begin{minipage}[t]{0.45\textwidth}
		\includegraphics[width=\textwidth]{pictures/results/a3c_vanilla/empty.png}
		
				\vspace*{-0.30cm}
\caption{Actor: Guided backpropagation. Gredients so small that you can't see them without a strong multiplication factor.}
\label{fig:break_a3c_act_guid_back}
	\end{minipage}
	\hfill
	\begin{minipage}[t]{0.45\textwidth}
		\includegraphics[width=\textwidth]{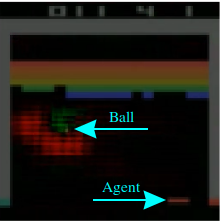}
		
				\vspace*{-0.30cm}
\caption{Critc: Guided backpropagation. Positive gradients are formed around the ball followed by negative gradients.}
\label{fig:break_a3c_crit_guid_back}
	\end{minipage}
\end{figure}
\begin{figure}[H]
	\begin{minipage}[t]{0.45\textwidth}
		\includegraphics[width=\textwidth]{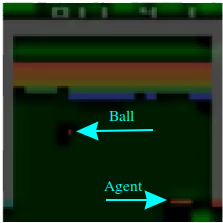}
			
				\vspace*{-0.30cm}	
\caption{Actor: Grad-Cam. Inverted gradients. Around the ball and the agent you can see that the neural net recognizes the agent and the ball (invented).  This is the only visualization method on the actor side that recognizes both the ball and the agent.}
\label{fig:break_a3c_act_grad_cam}
	\end{minipage}
\hfill
	\begin{minipage}[t]{0.45\textwidth}
		\includegraphics[width=\textwidth]{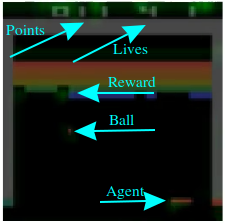}
		
				\vspace*{-0.30cm}
\caption{Critic: Grad-Cam. The ball and the agent are well highlighted by this visualization technique.}
\label{fig:break_a3c_crit_grad_cam}
	\end{minipage}
\end{figure}

\begin{figure}[H]
	\begin{minipage}[t]{0.45\textwidth}
		\includegraphics[width=\textwidth]{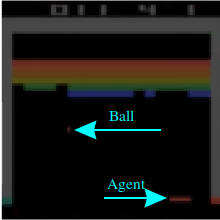}
		
				\vspace*{-0.30cm}
\caption{Actor: Guided Grad-Cam. Due to the inverted results of the Grad-Cam method no results will be shown here.}
\label{fig:break_a3c_act_guid_grad_cam}
	\end{minipage}
	\hfill
	\begin{minipage}[t]{0.45\textwidth}
		\includegraphics[width=\textwidth]{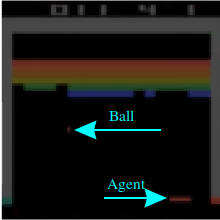}
		
				\vspace*{-0.30cm}
\caption{Critic: Guided Grad-Cam}
\label{fig:break_a3c_crit_guid_grad_cam}
	\end{minipage}
\end{figure}

\begin{figure}[H]
	\begin{minipage}[t]{0.45\textwidth}
		\includegraphics[width=\textwidth]{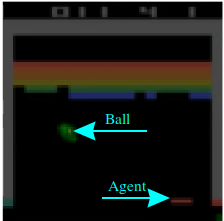}
		
				\vspace*{-0.30cm}
\caption{Actor: G1grad-Cam. This method shows us a very strong highlighting of the ball which is also very stable. However, only the ball is highlighted and not the agent.}
\label{fig:break_a3c_act_guid_g1grad_cam}
	\end{minipage}
	\hfill
	\begin{minipage}[t]{0.45\textwidth}
		\includegraphics[width=\textwidth]{pictures/results/a3c_vanilla/g1gradCam_actor_01.png}
		
				\vspace*{-0.30cm}
\caption{Critic: G1grad-Cam. This method shows us a very strong highlighting of the ball which is also very stable. However, only the ball is highlighted and not the agent.}
\label{fig:break_a3c_crit_g1guid_grad_cam}
	\end{minipage}
\end{figure}

\begin{figure}[H]
	\begin{minipage}[t]{0.45\textwidth}
		\includegraphics[width=\textwidth]{pictures/results/a3c_vanilla/empty.png}
		
				\vspace*{-0.30cm}
\caption{Actor: G2grad-Cam. As we have not seen any gradients in the Guided Backpropagation method we do not see any highlighting here either.}
\label{fig:break_a3c_act_guid_g2grad_cam}
	\end{minipage}
	\hfill
	\begin{minipage}[t]{0.45\textwidth}
		\includegraphics[width=\textwidth]{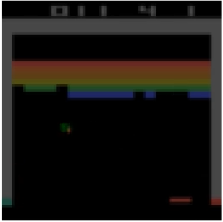}
		
				\vspace*{-0.30cm}
\caption{Critic: G2grad-Cam. Slight gradients can be seen on the ball.}
\label{fig:break_a3c_crit_g2guid_grad_cam}
	\end{minipage}
\end{figure}

%----------------------------------------250----
\begin{figure}[H]
	\begin{minipage}[t]{0.45\textwidth}
		\includegraphics[width=\textwidth]{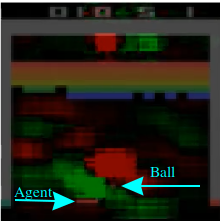}
		
				\vspace*{-0.30cm}
\caption{Actor: Gradient. Visualization of the actor gradient with a multiplication of 250.}
\label{fig:break_a3c_act_grad2}
	\end{minipage}
	\hfill
	\begin{minipage}[t]{0.45\textwidth}
		\includegraphics[width=\textwidth]{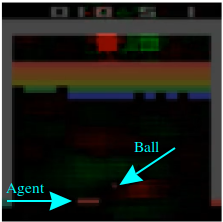}
		
				\vspace*{-0.30cm}
\caption{Critic: Gradient.}
\label{fig:break_a3c_crit_grad2}
	\end{minipage}
\end{figure}
\begin{figure}[H]

	\begin{minipage}[t]{0.45\textwidth}
		\includegraphics[width=\textwidth]{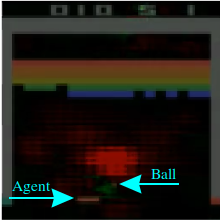}
		
				\vspace*{-0.30cm}
\caption{Actor: Guided backpropagation. Visualization of the actor gradient with a multiplication of 250.}
\label{fig:break_a3c_act_guided_back2}
	\end{minipage}
	\hfill
	\begin{minipage}[t]{0.45\textwidth}
		\includegraphics[width=\textwidth]{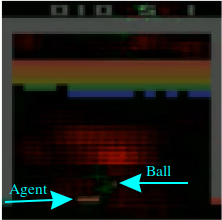}
		
				\vspace*{-0.30cm}
\caption{Critic: Guided backpropagation}
\label{fig:break_a3c_crit_guided_back2}
	\end{minipage}
\end{figure}

%------------------------------------------ A3C: Seaquest-v0 ------------------------------------------
\subsection{A3C: Seaquest-v0}

\begin{figure}[H]
	\begin{minipage}[t]{0.45\textwidth}
		\includegraphics[width=\textwidth]{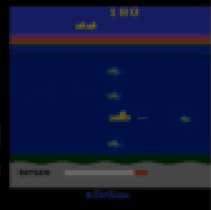}
		
		\vspace*{-0.30cm}
\caption{Actor: Gradient. Too small gradients to visualize them just like in the game Breakout-v0.}
\label{fig:a3c_act_grad}
	\end{minipage}
	\hfill
	\begin{minipage}[t]{0.45\textwidth}
		\includegraphics[width=\textwidth]{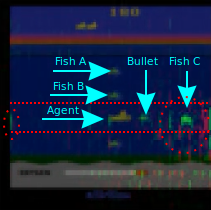}
		
		\vspace*{-0.30cm}
\caption{Critic: Gradient. Here you can see very clearly that the agent has learned to turn and shoot only left and right and never leaves the red area. You can see that the fish in this area are very strongly highlighted and outside this area there are almost no gradients to be seen. Sometimes gradients also appear on the left and right sides of the agent, where no fish can be seen (red circle).}
\label{fig:a3c_crit_grad}
	\end{minipage}
\end{figure}
\begin{figure}[H]
	\begin{minipage}[t]{0.45\textwidth}
		\includegraphics[width=\textwidth]{pictures/results/a3c_vanilla/seaquest/Seaquest_intro_A3C_Actor_grad_and_guid.png}
		
		\vspace*{-0.30cm}
\caption{Actor: Guided backpropagation. Too small gradients to visualize them just like in the game Breakout-v0.}
\label{fig:a3c_act_guid_back}
	\end{minipage}
	\hfill
	\begin{minipage}[t]{0.45\textwidth}
		\includegraphics[width=\textwidth]{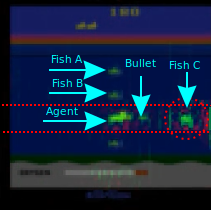}
		
		\vspace*{-0.30cm}
\caption{Critic: Guided backpropagation. Here you can see very clearly that the agent has learned to turn and shoot only left and right and never leaves the red area. You can see that the fish in this area are very strongly highlighted and also the agent itself. Outside this area there are almost no gradients to be seen.}
\label{fig:a3c_crit_guid_back}
	\end{minipage}
\end{figure}
\begin{figure}[H]
	\begin{minipage}[t]{0.45\textwidth}
		\includegraphics[width=\textwidth]{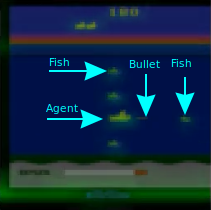}
		
		\vspace*{-0.30cm}
\caption{Actor: Grad-Cam. The fish and the agent itself are clearly highlighted. This method gives better results for the actor as well as for the Speil Breakout-v0 than the gradient methods, where the gradients are too small to visualize them without a multiplication factor.}
\label{fig:a3c_act_grad_cam}
	\end{minipage}
	\hfill
	\begin{minipage}[t]{0.45\textwidth}
		\includegraphics[width=\textwidth]{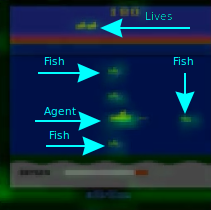}
		
		\vspace*{-0.30cm}
\caption{Critic: Grad-Cam. The fish and the agent itself are clearly highlighted. The fish and the agent itself are clearly highlighted.}
\label{fig:a3c_crit_grad_cam}
	\end{minipage}
\end{figure}
\begin{figure}[H]
	\begin{minipage}[t]{0.45\textwidth}
		\includegraphics[width=\textwidth]{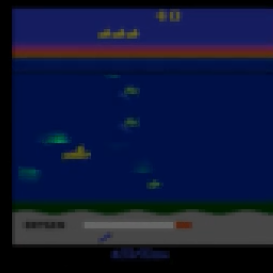}
		
		\vspace*{-0.30cm}
\caption{Actor: G1Grad-Cam. Close to the important features some gradients are visualized.}
\label{fig:a3c_act_g1grad_cam}
	\end{minipage}
	\hfill
	\begin{minipage}[t]{0.45\textwidth}
		\includegraphics[width=\textwidth]{pictures/results/a3c_vanilla/seaquest/Seaquest_intro_A3C_Actor_Critic_g1grad_cam.png}
		
		\vspace*{-0.30cm}
\caption{Critic: G1Grad-Cam. Close to the important features some gradients are visualized.}
\label{fig:a3c_crit_g1grad_cam}
	\end{minipage}
	
%--------------------------------------- A3C with LSTM: Seaquest-v0 --------------------------------------

\subsection{A3C with LSTM: Seaquest-v0}

\end{figure}
\begin{figure}[H]
	\begin{minipage}[t]{0.45\textwidth}
		\includegraphics[width=\textwidth]{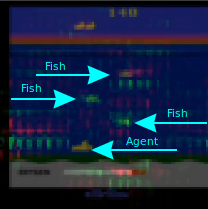}
		
		\vspace*{-0.30cm}
\caption{Actor: Gradient}
\label{fig:a3cLSTM_a_grad}
	\end{minipage}
	\hfill
	\begin{minipage}[t]{0.45\textwidth}
		\includegraphics[width=\textwidth]{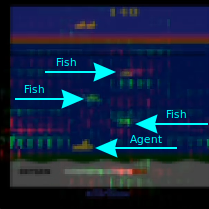}
		
		\vspace*{-0.30cm}
\caption{Critic: Gradient}
\label{fig:a3cLSTM_c_grad}
	\end{minipage}
\end{figure}
%--------------------------
\begin{figure}[H]
	\begin{minipage}[t]{0.45\textwidth}
		\includegraphics[width=\textwidth]{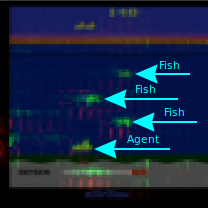}
		
		\vspace*{-0.30cm}
\caption{Actor: guided backpropagation}
\label{fig:a3cLSTM_a_guid_back}
	\end{minipage}
	\hfill
	\begin{minipage}[t]{0.45\textwidth}
		\includegraphics[width=\textwidth]{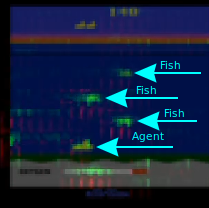}
		
		\vspace*{-0.30cm}
\caption{Critic: guided backpropagation}
\label{fig:a3cLSTM_c_guid_back}
	\end{minipage}
\end{figure}
%--------------------------
\begin{figure}[H]
	\begin{minipage}[t]{0.45\textwidth}
		\includegraphics[width=\textwidth]{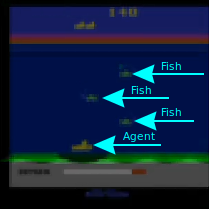}
		
		\vspace*{-0.30cm}
\caption{Actor: Grad-Cam}
\label{fig:a3cLSTM_a_grad_cam}
	\end{minipage}
	\hfill
	\begin{minipage}[t]{0.45\textwidth}
		\includegraphics[width=\textwidth]{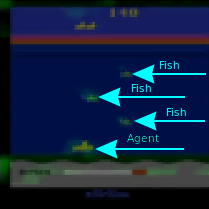}
		
		\vspace*{-0.30cm}
\caption{Critic: Grad-Cam}
\label{fig:a3cLSTM_c_grad_cam}
	\end{minipage}
\end{figure}
\end{document}